# Detection of Myocardial Infarction Based on Novel Deep Transfer Learning Methods for Urban Healthcare in Smart Cities

Ahmed Alghamdi[1], Mohamed Hammad[2], Hassan Ugail[3], Asmaa Abdel-Raheem[4], Khan Muhammad[5], Hany S. Khalifa[6], Ahmed A. Abd El-Latif [7,8,9]

[1] Department of Cybersecurity, College of Computer Science and Engineering, University of Jeddah, Saudi Arabia

[2] Information Technology Department, Faculty of Computers and Information, Menoufia University, Egypt

[3] Centre for Visual Computing, University of Bradford, Bradford BD7 1DP, UK

[4] Public health and Community Medicine Department, Faculty of Medicine, Menoufia University, Shebin El-Koom, Egypt

[5] Intelligent Media Lab, Department of Software, Sejong University, Seoul, Republic of Korea

[6] Computer science department, Misr Higher Institute of Commerce and Computers, Egypt

[7] Mathematics and Computer Science Department, Faculty of Science, Menoufia University, Shebin El-Koom, Egypt

[8] School of Information Technology and Computer Science, Nile University, Egypt

[9] School of Computer Science and Technology, Harbin Institute of Technology, Harbin 150080, China

**Abstract:** Myocardial infarction (MI), also known as a cardiac attack, is one of the common cardiac disorders occurs when one or more coronary arteries are blocked. Early detection of MI is critical for the reduction of the rising death rate. Cardiologists use electrocardiogram (ECG) as a diagnostic tool to monitor and reveal the MI signals. However, all the MI signals are not constant and noisy, so it is tough to detect or observe these signals manually. Manually studying large amounts of ECG data can be tedious and time-consuming. Therefore, there is a need for methods to automatically analyze the ECG data and make diagnosis. Number of studies has been presented to address MI detection, but most of these methods are computationally expensive and faces the problem of overfitting while dealing real data. In this paper, an effective computer-aided diagnosis (CAD) system is presented to detect MI signals using the convolution neural network (CNN) for urban healthcare in smart cities. Two types of transfer learning techniques are employed to retrain the pre-trained VGG-Net (Fine-tuning and VGG-Net as fixed feature extractor) and obtained two new networks VGG-MI1 and VGG-MI2. In the VGG-MI1 model, the last layer of the VGG-Net model is replaced with a specific layer according to our requirements and various functions are optimized to reduce overfitting. In the VGG-MI2 model, one layer of the VGG-Net model is selected as a feature descriptor of the ECG images to describe it with informative features. Considering the limited availability of dataset, ECG data is augmented which has increased the classification performance. Physikalisch-technische bundesanstalt (PTB) Diagnostic ECG database is used for experimentation, which has been widely employed in MI detection studies. In case of using VGG-MI1, we achieved an accuracy, sensitivity, and specificity of 99.02%, 98.76%, and 99.17%, respectively and we achieved an accuracy of 99.22%, a sensitivity of 99.15%, and a specificity of 99.49% with VGG-MI2 model. Experimental results validate the efficiency of the proposed system in terms of accuracy sensitivity, and specificity.

## 1. Introduction

According to the world health organization (WHO), coronary heart disease, also known as ischemic heart disease, is the lead cause of deaths in the world. Over 17.7 million people die annually from cardiovascular diseases (CVDs), and over 70% of these deaths are due to heart attacks [1]. Partial or complete occlusion can cause inadequate blood flow toward coronary arteries, leading to the myocardial ischemia [2]. Myocardial infarction (MI), usually identified as a heart attack, is one of the common cardiac disorders caused by a prolonged myocardial ischemia [3]. MI is a serious result of coronary artery disease. The heart functionality highly depends on the coronary circulation system that supplies oxygenated blood directly toward cardiac muscles to keep the heart nourished and oxygenated. MI occurs when a coronary artery is so severely blocked leading to the insufficient supply of nutrients and the oxygen-rich blood to a section of heart muscle. Sudden death occurs within an hour of the beginning indications of this disease [4]. Thus, it is important to periodically monitor the heart rhythms to manage and prevent the MI and decrease the risk of the subject's death. Several methods including electrocardiogram (ECG), which is the most common medical tool that provides information about the waveform of the heartbeat can diagnose patients with MI. Other common diagnostic tests include magnetic resonance imaging (MRI) [5] and echocardiography [6]. However, the ECG is the first MI diagnostic method for urgent patients because it is easy to use and cost-effectively. Analyzing the MI signals manually is slightly hard because the nature of all the MI signals is not constant and noisy. Thus, numerous computer-aided diagnosis systems (CADs) have been proceeded to solve these difficulties. Nowadays, machine learning is a very common and effective tool to solve problems in the field of medicine [7-9], especially for MI detection [10-17].

Several studies have been carried out to automate MI detection by analyzing ECG signals [10-17]. Sadhukhan *et al.* [10] developed an automated ECG analysis algorithm based on the harmonic phase distribution pattern of the ECG data for MI identification. They yielded an average detection accuracy of 95.6% with a sensitivity of 96.5% and specificity of 92.7%. Jayachandran *et al.* [11] used the multi-resolution properties of wavelet transform to analyze the normal and MI ECG signals with an accuracy of 96.1%. Dohare *et al.* [12] detected the MI signals using 12-lead ECG data. They achieved a sensitivity of 96.66%, specificity of 96.66% and an accuracy of 96.66% with support vector machine (SVM) classifier. Arif *et al.* [13] implemented k-nearest neighbor (kNN) classifier and extracted time domain features of ECG beats to detect and localize the MI signals. They achieved a sensitivity of 99.97% and specificity of 99.9% for MI detection. However, in these studies [10,11,12,13], the authors adopted on the classic machine learning approaches, which often suffer from overfitting and performance degradation when validated on a separate dataset. In this study, we did not follow these classical approaches, but build the proposed system with a convolutional neural network (CNN), which is one of the most widely used deep learning models. Recently, CNN has been utilized in the automated detection of abnormal heart conditions [18-24].

There are several approaches have been also proposed to detect MI by digital analysis of ECG signals [14-17]. However, few of previous works based on CNN have been utilized for MI detection [25,26,27]. Wu *et al.* [25] employed deep feature learning and soft-max regression as a multi-class classifier for MI detection. They also incorporated multi-scale discrete wavelet transform into the feature learning process to increase the feature learning process. Their method yielded a sensitivity of 99.64% and specificity of 99.82%. Acharya *et al.* [26] detected the MI signals using an 11-layer 1-D CNN. They achieved an average accuracy of 95.22% using noise filters and 93.53% without using any noise filters. Liu *et al.* [27] proposed a method called multiple-feature-branch CNN (MFB-CNN) for MI detection with an accuracy of 98.79%. However, these works [25,26,27] used big convolution filters, which lead to increase the computation cost. Moreover, a few ECG segments were analyzed in most of these works and hence resulted in low specificity and sensitivity.

Following are the major shortcomings of the existing machine learning and deep learning-based MI detection approaches:

- Computationally complex algorithms with long authentication time.
- Sensitive to the quality of ECG signals.
- Work on multiple leads.
- Intensive to learn the features.
- Required large amount of ECG data.

- Suffer from overfitting and performance degradation when validated on other datasets except the training one.

Therefore, to overcome the previous limitations, this study proposes a MI detection method using deep two-dimensional CNN with grayscale ECG images for deep urban healthcare in smart cities. The proposed method aims to achieve high classification accuracy than existing approaches, with following main goals:

- Design a new MI detection algorithm, which achieves superior results compared with the previous hand engineered techniques.
- Design a light-weight deep learning-assisted MI detection framework which is suitable for real-time ECG analysis.
- Building a model that achieves high accuracy with a small number of training ECG records.
- Overcome the overfitting problem faced by previous works.

In this work, we have used two ways of the transfer learning technique [28]. The first way is fine-tuning the VGG-Net model and obtaining new network VGG-MI1. The second way is using the VGG-Net model as a feature extractor and obtaining new network VGG-MI2. Unlike most of the previous works, segmentation and feature extraction are no longer required in the proposed method. In addition, we optimized the proposed CNN model using data augmentation, which improve the accuracy of the proposed system by 2% in absolute values. Besides data augmentation, dropout technique is used [29] to avoid overfitting. The proposed algorithm is tested on the physikalisch-technische bundesanstalt (PTB) database [30] to evaluate its performance. Results show that the proposed algorithm achieves superior results compared with the previous algorithms based on CNN. Considering the drawbacks of previous studies, the main contributions of this paper can be summarized as follows:

- A pre-trained deep CNN model is presented for MI detection, where we use VGG-Net that designed for the object recognition tasks to achieve state-of-the-art accuracy in MI detection. Also, we use VGG-Net as a feature extractor by selecting valuable layers to get a good representation of ECG data.
- Unlike most of previous deep learning approaches, we employed 3×3 filter in the first convolutional layers to reduce the computing cost and the noise effect. Moreover, we employ small number of pooling layers, which make the proposed two models more stable when using the input ECG image with size 128×128.
- We propose a method that work on the original ECG signals without using any signal filtering, which makes our method insensitive to the ECG signal quality.

- The proposed model is trained using only one ECG lead signal unlike most of previous algorithms. Hence, it reduces the computing cost and makes our method less complex than other previous methods.
- Data augmentation is successfully used to increase the robustness of the proposed system against small variations, which achieved high specificity and sensitivity in the results.
- We employ Q-Gaussian multi-class support vector machine (QG-MSVM) classifier, which plays an important rule for increasing the accuracy of the proposed system and can solve the small sample problem.

The rest of this paper is structured as follows: Section 2 gives the proposed myocardial infarction detection framework for urban healthcare in smart cities. In section 3, the proposed deep transfer learning method for the detection myocardial infarction is introduced. The dataset used for the experiments of the proposed approach is given in Section 4. Experimental analysis and results are given in Section 5. Section 6 is devoted to the discussion of the results. Finally, Section 7 concludes this paper.

## 2. The proposed myocardial infarction detection framework for urban healthcare in smart cities

Nowadays, computer vision equipped with modern technologies such as deep learning has the potential to demonstrate its possibilities beyond the surveillance and law enforcement purposes. It is true that its first use has been mainly motivated by security reasons. From the autonomous vehicles to the interactive and intelligent architectures, groundbreaking applications open new ways to contribute to goals of smart cities concepts. These include a wide range of solutions in e-health care, intelligent transportation, surveillance, vehicle identification and tracking, smart parking, crowd density, and monitoring, etc. This paper aims to provide a platform to research community and professionals to demonstrate solutions and address research challenges in the development of detection of myocardial infarction using novel deep transfer learning methods for urban healthcare in smart cities.

Figure.1 shows the proposed myocardial infarction detection framework for urban healthcare in smart cities. The ECG signals are captured from medical sensors/devices in one side and send to the cloud-based E-healthcare for processing by the proposed deep transfer learning model in order to detect the myocardial infarction for patients. All the data are processing in urban healthcare in the smart city.

## 3. Methodology

In this study, the proposed method is validated on two ECG datasets with the same number of ECG beats. We removed the noise in one dataset using a band-pass (Butterworth) filter with 0.5Hz and 40Hz cutoff frequencies. In the second dataset, we kept the noise in the ECG signals. After that, we carried out the peaks (P, R, and T) detection using our previous algorithm [31]. We defined a two ECG beats image using the detected peaks while excluding the first and the last ECG beat. Finally, the classification is performed with these obtained ECG images in CNN classification stage.

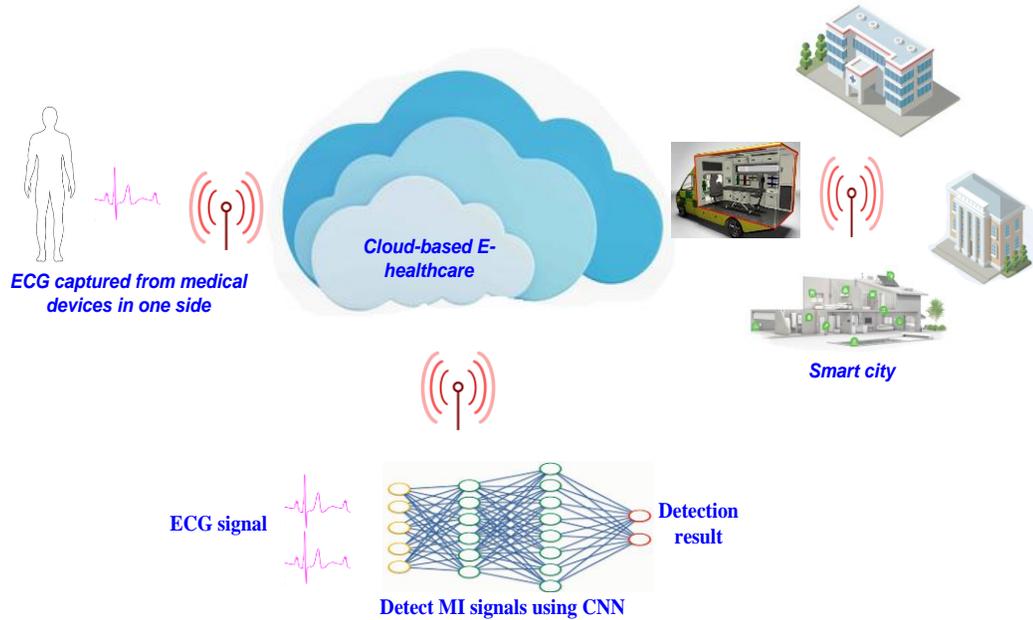

Fig. 1. The proposed myocardial infarction detection framework for urban healthcare in smart cities

### 3.1. Preprocessing stage

We transformed the one-dimensional ECG signal to two-dimensional ECG image by plotting each ECG signal as an individual 128×128 gray-scale image [32,33]. The reason for working with *two*-dimensional ECG images is that *two*-dimensional (2D) convolutional and pooling layers are more suitable for filtering the spatial locality of the ECG images. As a result, higher ECG classification accuracy can be obtained. Also, we can apply data augmentation to the trained ECG images to improve the classification accuracy, which is difficult to be applied on a one-dimensional ECG signal.

### 3.2. CNN classification stage

In this work, we adopted CNN as the ECG MI classifier. CNN is a fundamental deep learning tool with many hidden layers and parameters [34]. CNN has been widely used in many fields such as image processing [35], pattern recognition [36,37] and other kinds of cognitive tasks [38,39]. Moreover, it is employed as an automated diagnostic tool in the medical fields [40,41,42]. A typical CNN has composed of *three*-layer types: the convolutional layer, the fully-connected layer, and the pooling layer. The convolutional layer is used for the detection of MI signal features. The pooling layer also called down-sampling layer used to control the overfitting and for reducing the weights number. The fully-connected layer used for classification.

### 3.2.1. Models

Recently, numerous CNN models have been developed for large-scale image classification such as Caffe-Net [43], Alex-Net [44], and VGG-Net [45]. The proposed algorithm is based on the VGG-Net for classification because it has a much deeper architecture than other models, hence it can provide many informative features [46]. The VGG-Net outperforms the previous generation of CNN models, which is trained with the public ImageNet dataset and achieves the second place in the detection task of the ImageNet 2014 challenge [47].

### 3.2.2. The architecture of MI-CNN

We retrained the pre-trained VGG-Net and obtained two new networks VGG-MI1 and VGG-MI2. In both networks, the architecture of the VGG-Net is different from the standard VGG-Net by using small filter size in the first convolution layers and a small number of pooling layers, which lead to lower computation cost and makes the proposed models more stable for real-time ECG analysis. In addition to optimizing the standard VGG-Net, for VGG-MI1 we replaced the last 1000-unit SoftMax layer by a 2-unit SoftMax layer (shown in Figure 2), so the network can output the 2 classes (MI or normal) instead of the original 1000 classes that the network was designed for. In the second network VGG-MI2, the outputs of some selected layers are used as a feature descriptor of the input image to describe it by informative and significant features and we used external classifier (QG-MSVM) for the classification to speed up the training task and improve the classification accuracy.

Table 1 describes the architecture of the proposed 2D CNN model, which is a stack of convolutional and pooling layers followed by 3 fully-connected layers where the input images are resized to 128×128×1. The max-pooling layer, also known as a down-sampling layer, was used to reduce the computation complexity and control overfitting. The proposed CNN model does not contain Local Response Normalization (LRN), because it leads to an increase in computing time and cost and does not affect the results. We introduced non-linearity in the model by providing all layers with the Rectified Linear Unit (ReLU) activation function. In comparison with commonly used sigmoid and hyperbolic tan activation functions, ReLU does not saturate, thereby speeding up the convergence of stochastic gradient descent. The mathematical form of ReLU function is as follows:

$$f(x) = \max(0, x) \qquad (1)$$

Where *x* is the input to a neuron. We employed 3×3 filters in the first convolutional layers to reduce the computing cost and the noise effect [48], which differs from most of the previous models that used large filters (e.g. 5×5 or 10×10 [49,50]). The soft-max layer is the last layer of the network which is used to train the CNN. In this work, the dropout technique [29] is performed during the training phase with a probability of 0.5 and placed its position in the fully-connected layer to avoid overfitting. We do not apply a dropout to convolutional layers because they have a smaller number of parameters compared to the number of activations, which make the nodes not adaption together.

Table 1. The architecture of the proposed VGG-MI

| Layers No. | Type  | Kernel size | No. Kernels  | Stride | Input size  |
|---|---|---|---|---|---|
| Layer 1  | Conv1 | 3×3 | 64 | 1 | 128×128×1 |
| Layer 2  | Conv2 | 3×3 | 64 | 1 | 128×128×64 |
| Layer 3  | Pool  | 2×2 |    | 2 | 128×128×64 |
| Layer 4  | Conv3 | 3×3 | 128 | 1 | 64×64×64 |
| Layer 5  | Conv4 | 3×3 | 128 | 1 | 64×64×128 |
| Layer 6  | Pool  | 2×2 |    | 2 | 64×64×128 |
| Layer 7  | Conv5 | 3×3 | 256 | 1 | 32×32×128 |
| Layer 8  | Conv6 | 3×3 | 256 | 1 | 32×32×256 |
| Layer 9  | Pool  | 2×2 |    | 2 | 32×32×256 |
| Layer 10 | Full  |     | 2048 |  | 16×16×256 |
| Layer 11 | Full  |     | 2048-dropout |  | 2048 |
| Layer 12 | Soft  |     | 2 |  | 2048 |

### 3.2.3. Optimized CNN architecture

We used two ways of the transfer learning technique [28] to retrain the VGG-Net and obtained two new networks VGG-MI1 and VGG-MI2. The two ways are as follows:

A. Replacing the last 1000-unit soft-max layer, which is originally designed to predict 1000 classes in VGG-Net model by a 2-unit soft-max layer, which assigns the two classes normal and MI. This network is called VGG-MI1 as shown in Figure 2.
B. Selecting one layer as a feature descriptor of the training and test the ECG images to describe it with informative features. Second output fully-connected layer of the pre-trained net as a feature descriptor. This network is called VGG-MI2 as shown in Figure 3.

For the VGG-MI1 model, besides replacing the last 1000-unit soft-max layer, we optimized various functions of the standard VGG-Net model to reduce overfitting and improve classification accuracy. For the VGG-MI2 model, we selected the second fully connected layer to describe it with informative features because deeper layers contain higher-level features and also; we selected the layer that achieves the highest accuracy comparing to other layers in the proposed model as shown in Figure 4. Finally, the proposed models are compared with standard VGG-Net and Alex-Net.

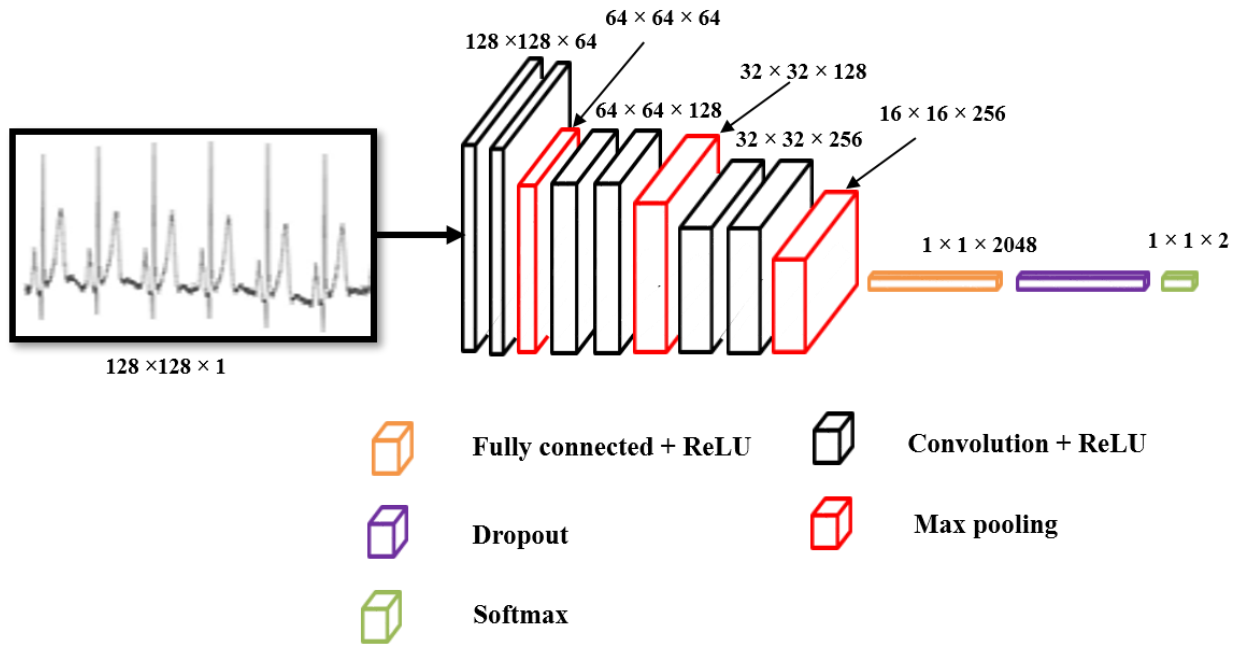

Fig. 2. The architecture of VGG-MI1

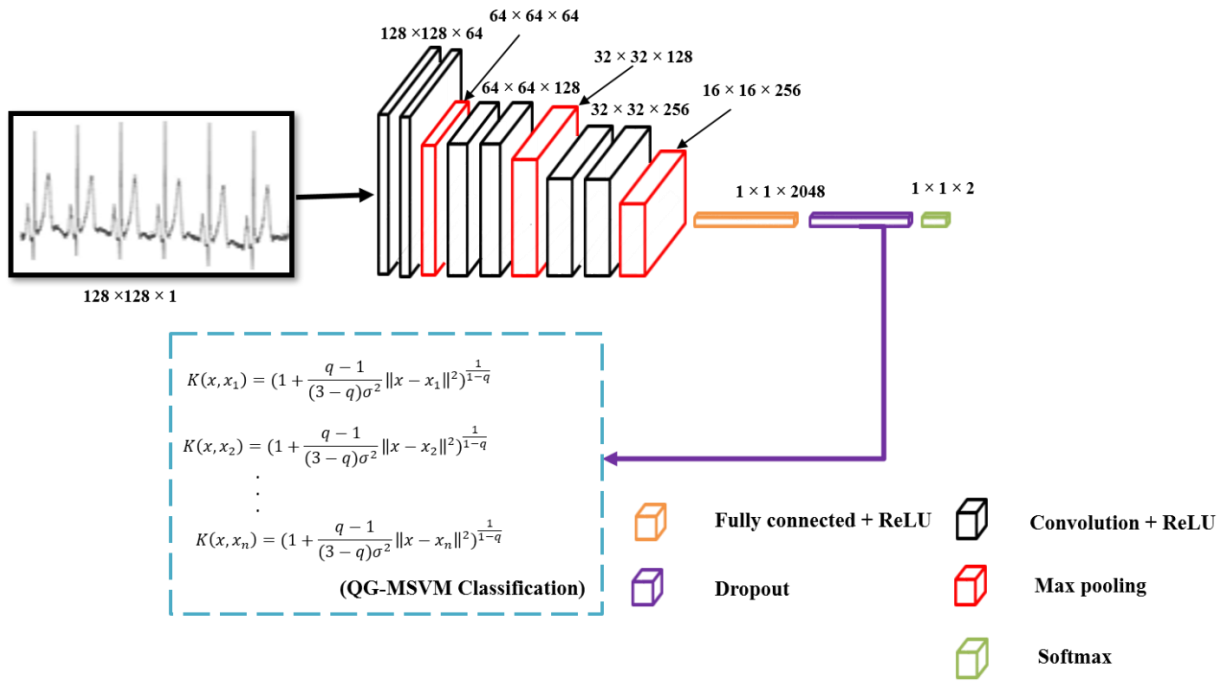

Fig. 3. The architecture of VGG-MI2

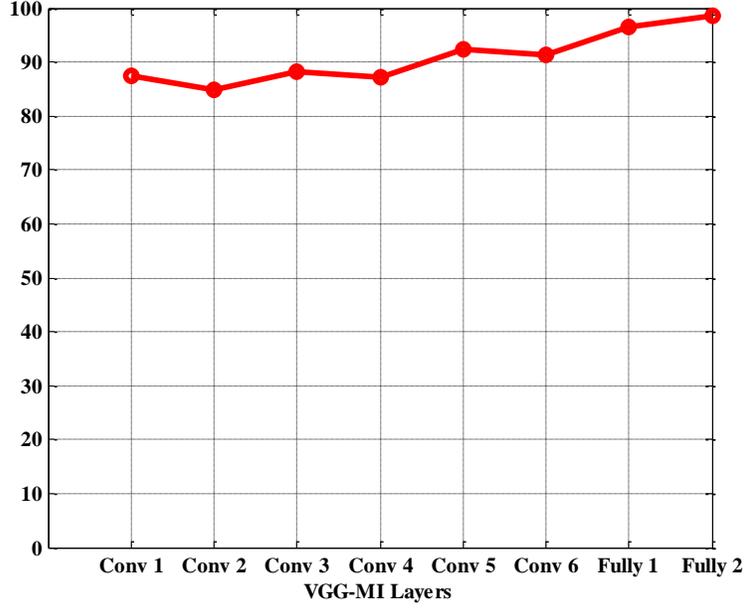

Fig. 4. Effect of the distinctive layer features on the classification accuracy

### 3.2.4. Training

The VGG-MI1 model was trained through stochastic gradient descent (SGD) with a minibatch size of five samples using momentum [51]. The weights of the CNN filters are initialized using random sampling from a Gaussian distribution with 0.01 standard deviation and zero means. The hyper-parameters are set as follows: learning rate of 0.001, weight decay of 0.0005 and the training and testing of the CNN were done in 50 epochs. These parameters are determined accordingly to obtain optimum performance. Table 2 shows the configuration of the proposed CNN in this work.

Table 2. The most important VGG parameters

| Parameter | Value |
| --- | --- |
| learning rate | 0.001 |
| standard deviation | 0.01 |
| mean | 0 |
| weight decay | 0.0005 |
| epochs | 50 |
| minibatch size | 5 samples |

The VGG-MI2 model was trained using Q-Gaussian multi-class support vector machine (QG-MSVM) classifier [52], which is defined as:

$$K(x, x_i) = (1 + \frac{q-1}{(3-q)\sigma^2} \|x - x_i\|^2)^{\frac{1}{1-q}} \qquad (2)$$

Where $q$ is a real-valued parameter, $\sigma$ is a real value standard variance of Gaussian distribution and each $x_i \in R_p$ is a p-dimensional real vector. In our previous work [52], we used QG-MSVM to classify fingerprints, and we achieved a good result comparing to other SVM kernels. In this work, we modified the QG-MSVM to classify ECG. We use the same values in our previous work [52], which gives the best results compared to other SVM kernels: $\frac{1}{\sigma^2}$ is assigned to 0.5 and *q* to 1.5.

### 3.2.5. Testing

We perform a test on the CNN modal after every completed round of training epoch. The data were separated into three parts: 60% of the data for training, 30% for validation and 10% for testing. Moreover, we employed a ten-fold cross-validation approach [53], then the performance of the system is evaluated in each fold and the average of all ten folds was calculated as the final performance of the system.

### 3.3. Data Augmentation

Data augmentation is a technique to generate artificial data samples from the original ones. It is used to make the proposed model more robust for overfitting. It has been successfully used in previous works in the medical analysis [54,55]. In this study, data is augmented as: From each image of the dataset nine smaller images with 75% of each dimension of the original images are extracted: four images cropped from each corner, one at the center, one at the right center, one at the left center, one at the top center and one at the bottom center. After that, these augmented images are resized to the original image size 128×128. Therefore, we obtained a database that is 10 times larger than the original one. Figure 5 shows nine example cropped images with the original image.

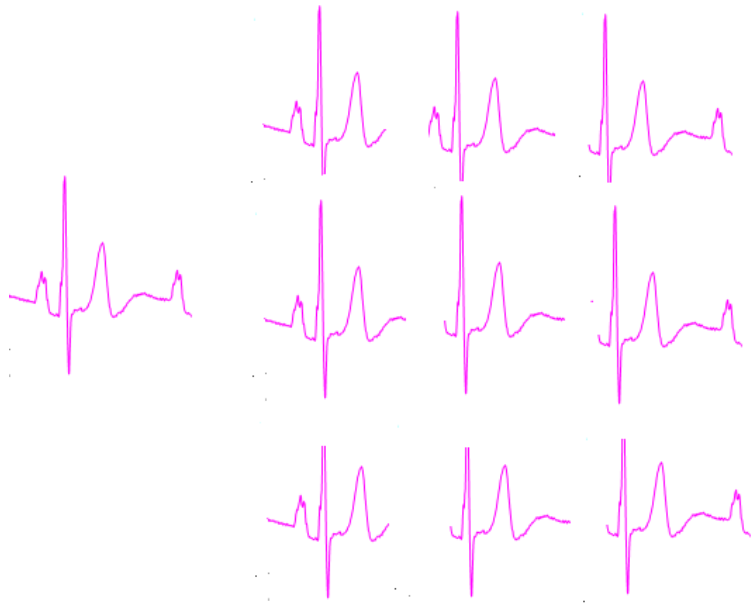

Fig. 5. Original ECG image and nine cropped images

## 4. Dataset

The PTB database [30] was used in this paper, which has been widely employed in MI detection studies [10-17,25,26,27]. This database [30] contains 549 records from 290 subjects and each subject is represented by one to five records. Overall, the age range of the participants was between 17 and 87 (28% were female and 72% were male). It contains 549 records such that each record includes 15 simultaneously measured signals; the conventional 12 leads together with the three Frank lead ECG. The ECG signals are digitized at 1000 samples per second (1000Hz). 147 out of 290 subjects and 368 out of 549 records are labelled as MI cases. In this work, we have used two-second duration of Lead II ECG signals with a total of 21,092 normal ECG beats and 80,364 MI ECG beats, where each two-beat represent a 128×128 gray-scale image. Figure 6 shows an illustration of two seconds of normal and MI ECG signals with and without noise.

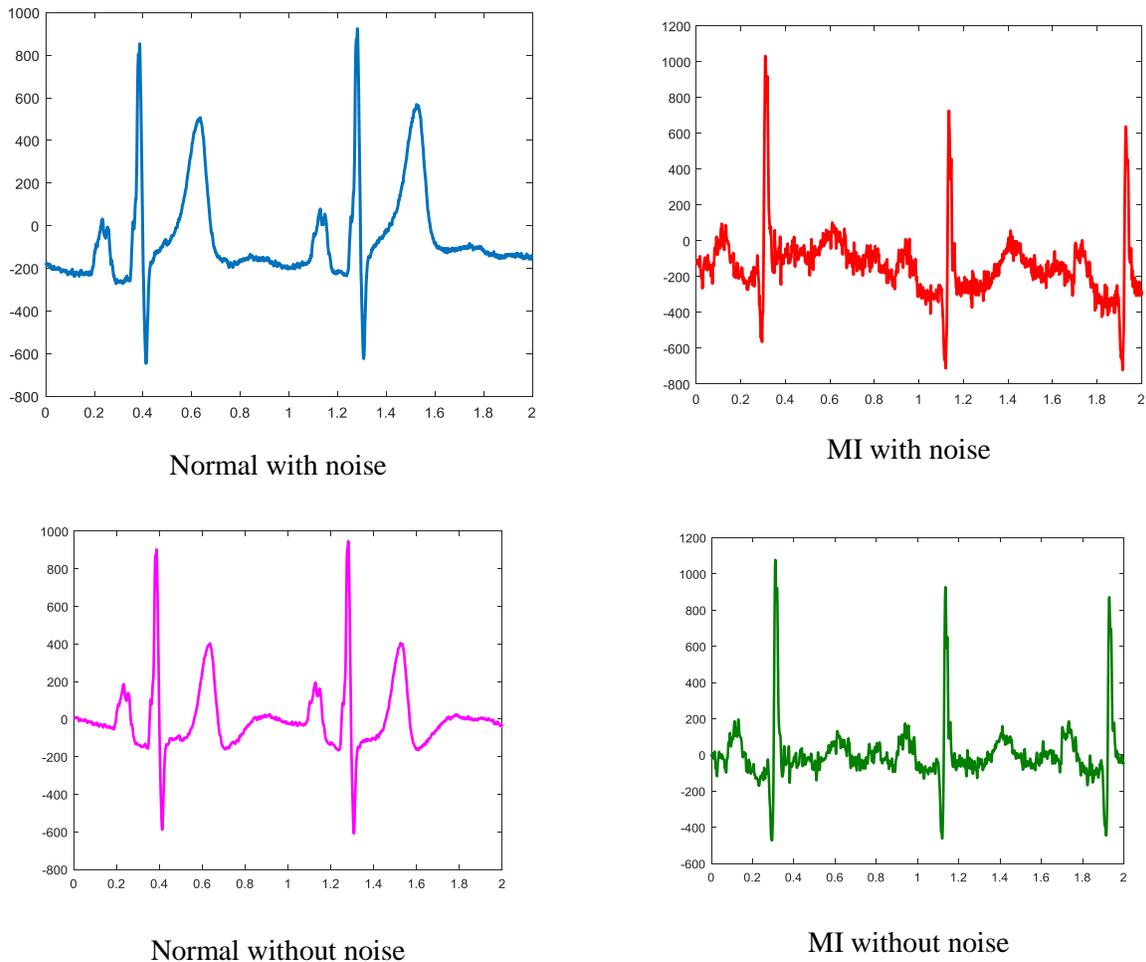

Fig. 6. Sample Normal and MI ECG signal with and without noise

## 5. Experimental Results

The proposed algorithm is evaluated on a PC workstation with 2.7-GHz CPU, 32GB of memory and a moderate GPU card. All methods have been executed using MATLAB R2017a on Microsoft Windows 10 Pro 64-bit. The performance of the proposed two CNN models was compared with two well-known CNN models, VGG-Net and Alex-Net. As well, we compare the performance of the proposed models with a different approach based on deep learning for MI detection. The PTB database is used for the evaluation of the experiments.

### 5.1. Performance metrics

To assess the performance of the proposed algorithm several parameters are used:

1. The Sensitivity (Se) is defined as (3):

$$\text{Se}(\%) = \frac{TP}{FN+TP} \times 100 \qquad (3)$$

2. The Predictivity (Pre) is defined as (4):

$$\text{Pre}(\%) = \frac{TP}{TP+FP} \times 100 \qquad (4)$$

3. The Specificity (Spe) is defined as (5):

$$\text{Spe}(\%) = \frac{TN}{TN+FP} \times 100 \qquad (5)$$

4. The Accuracy (Acc) is defined as (6):

$$\text{Acc}(\%) = \frac{TP+TN}{TP+TN+FP+FN} \times 100 \qquad (6)$$

where:

The number of MI signals that classifies as normal is the False Negative (FN), False Positive (FP) denotes the number of normal ECG signals that classifies as MI, the number of normal ECG signals that classifies as normal is the True Negative (TN) and True Positive (TP) is the number of MI signals that classifies as MI.

### 5.2 Results

Four scenarios are used in this paper, the first scenario is the best-case1, in this case, we worked on MI signals without noise and we employed the augmentation technique on the trained data. The second scenario is the best-case2, in this case, we worked on MI signals without noise and without using augmentation technique. The third scenario is the worst-case1, in this case, we worked on MI signals with noise and using augmentation technique on the trained data. The last scenario is the worst-case2, in this case, we worked on MI signals with noise and without using augmentation technique. Acc, Se, Pre and Spe are calculated in all scenarios and the comparison is done between the proposed two models (VGG-MI1 and VGG-MI2) and the standard VGG-Net and Alex-Net on the PTB database.

The confusion matrix of the results for the two MI detection models with and without augmentation techniques across all folds is presented in Tables 3, 4, 5 and 6 respectively. It can be noted from the Table 3 that, 97.53% of ECG segments are correctly classified as Normal class, 96.95% of ECG segments are correctly classified as MI when using VGG-MI1 on ECG segments without noise and without augmentation technique and 99.17% of ECG segments are correctly classified as Normal class, 98.76% of ECG segments are correctly classified as MI when using augmentation technique. Also, in Table 4, 93.59% of ECG segments are correctly classified as Normal class, 92.53% of ECG segments are correctly classified as MI when using VGG-MI1 on ECG segments with noise and without augmentation technique and 94.83% of ECG segments are correctly classified as Normal class, and 95.46% of ECG segments are correctly classified as MI when using augmentation technique. For VGG-MI2, it can be noted from the Table 5 that, 98.53% of ECG segments are correctly classified as Normal class, 97.95% of ECG segments are correctly classified as MI when using VGG-MI2 on the second scenario (best-case2) and 99.49% of ECG segments are correctly classified as Normal class, 99.15% of ECG segments are correctly classified as MI when working on the first scenario (best-case1). Also, in Table 6, 95.54% of ECG segments are correctly classified as Normal class, 95.08% of ECG segments are correctly classified as MI when using VGG-MI2 on the fourth scenario (worst-case2) and 96.79% of ECG segments are correctly classified as Normal class, and 97.35% of ECG segments are correctly classified as MI when working on the third scenario (worst-case1).

For VGG-MI1, a total of 2.4% Normal ECG segments are wrongly classified as MI and a total of 2.3% MI segments are wrongly classified as Normal without using augmentation technique after removing the noise. A total of 0.8% Normal ECG segments are wrongly classified as MI and a total of 1% MI segments are wrongly classified as Normal using augmentation technique after removing the noise.

For VGG-MI2, a total of 1.4% Normal ECG segments are wrongly classified as MI and a total of 2% MI segments are wrongly classified as Normal without using augmentation technique after removing the noise. A total of 0.5% Normal ECG segments are wrongly classified as MI and a total of 0.8% MI segments are wrongly classified as Normal using augmentation technique after removing the noise.

Table 3. Confusion matrix of ECG segments **without noise** across 10-fold for **VGG-MI1**

| VGG-MI1 | **True/Predicted** | **Normal** | **MI** | **Acc (%)** | **Se (%)** | **Pre (%)** | **Spe (%)** |
|---|---|---|---|---|---|---|---|
| **Without augmentation** | **Normal** | 20573 | 519 | 97.57 | 97.53 | 91.36 | 96.95 |
| | **MI** | 1945 | 78419 | 97.57 | 96.95 | 99.34 | 97.53 |
| **With augmentation** | **Normal** | 209174 | 1746 | 99.02 | 99.17 | 96.24 | 98.76 |
| | **MI** | 8168 | 795472 | 99.02 | 98.76 | 99.78 | 99.17 |

Table 4. Confusion matrix of ECG segments **with noise** across 10-fold for **VGG-MI1**

| VGG-MI1 | True/Predicted | Normal | MI | Acc (%) | Se (%) | Pre (%) | Spe (%) |
|---|---|---|---|---|---|---|---|
| Without augmentation | Normal | 19744 | 1352 | 92.75 | 93.59 | 76.68 | 92.53 |
| Without augmentation | MI | 6003 | 74361 | 92.75 | 92.53 | 98.21 | 93.59 |
| With augmentation | Normal | 200031 | 10889 | 95.33 | 94.83 | 84.59 | 95.46 |
| With augmentation | MI | 36421 | 767219 | 95.33 | 95.46 | 98.60 | 94.83 |

Table 5. Confusion matrix of ECG segments **without noise** across 10-fold for **VGG-MI2**

| VGG-MI2 | True/Predicted | Normal | MI | Acc (%) | Se (%) | Pre (%) | Spe (%) |
|---|---|---|---|---|---|---|---|
| Without augmentation | Normal | 20783 | 309 | 98.07 | 98.53 | 92.68 | 97.95 |
| Without augmentation | MI | 1640 | 78724 | 98.07 | 97.95 | 99.60 | 98.53 |
| With augmentation | Normal | 209856 | 1064 | 99.22 | 99.49 | 96.84 | 99.15 |
| With augmentation | MI | 6828 | 796812 | 99.22 | 99.15 | 99.86 | 99.49 |

Table 6. Confusion matrix of ECG segments **with noise** across 10-fold for **VGG-MI2**

| VGG-MI2 | True/Predicted | Normal | MI | Acc (%) | Se (%) | Pre (%) | Spe (%) |
|---|---|---|---|---|---|---|---|
| Without augmentation | Normal | 20152 | 940 | 95.17 | 95.54 | 83.60 | 95.08 |
| Without augmentation | MI | 3951 | 76413 | 95.17 | 95.08 | 98.78 | 95.54 |
| With augmentation | Normal | 204170 | 6750 | 97.24 | 96.79 | 90.64 | 97.35 |
| With augmentation | MI | 21250 | 782390 | 97.24 | 97.35 | 99.14 | 96.79 |

From previous Tables, we can show that the best results are obtained when using the first two scenarios (best-case1 and best-case2) so; we worked on these two scenarios in all comparisons in this paper.

The overall classification results for the proposed two MI detection models using the first and the second scenarios are collected in Tables 7 and 8.

Table 7. The overall classification results for **VGG-MI1** using the first and the second scenarios

| VGG-MI1 | TP | TN | FP | FN | Acc (%) | Se (%) | Pre (%) | Spe (%) |
|---|---|---|---|---|---|---|---|---|
| With Augmentation | 795472 | 209174 | 1746 | 8168 | 99.02 | 98.76 | 99.78 | 99.17 |
| Without Augmentation | 78149 | 20573 | 519 | 1945 | 97.57 | 96.95 | 99.34 | 97.53 |

Table 8. The overall classification results for **VGG-MI2** using the first and the second scenarios

| VGG-MI2 | TP | TN | FP | FN | Acc (%) | Se (%) | Pre (%) | Spe (%) |
|---|---|---|---|---|---|---|---|---|
| With Augmentation | 796812 | 209856 | 1064 | 6828 | 99.22 | 99.15 | 99.86 | 99.49 |
| **Without Augmentation** | 78724 | 20783 | 309 | 1640 | 98.07 | 97.95 | 99.60 | 98.53 |

It can be seen from Tables 7 and 8 that: an accuracy of 99.02% and a sensitivity and specificity of 98.76% and 99.17% respectively are achieved using the first MI detection model (VGG-MI1) using augmentation after removing the noise. Also, an average accuracy, sensitivity, and specificity of 99.22%, 99.15%, and 99.49% respectively are obtained for the second MI detection model (VGG-MI2) using augmentation after removing the noise. The performance of the VGG-MI2 is better than the VGG-MI1 because we employed external classifier (the QG-MSVM classifier) in the VGG-MI2 model, which plays an important rule for increasing the accuracy of the proposed system. As well, QG-MSVM can solve the small sample problem that faces the CNN training, which usually requires large sample data. Thus, the performance of VGG-MI2 when using augmentation and without augmentation is better than VGG-MI1. Table 9 summarizes a comparative study of the proposed two MI detection models with standard VGG-Net and Alex-Net.

Table 9. Summarizes evaluation results of the proposed models and other two models using the first and the second scenarios (the results for the proposed approaches are given in **bold**)

| Model | Acc (%) | Se (%) | Pre (%) | Spe (%) |
|---|---|---|---|---|
| VGG-Net with Augmentation | 97.35 | 97.44 | 99.20 | 97.02 |
| VGG-Net without Augmentation | 97.11 | 97.25 | 99.07 | 96.54 |
| Alex-Net with Augmentation | 98.69 | 98.63 | 99.67 | 98.77 |
| Alex-Net without Augmentation | 98.24 | 98.03 | 99.73 | 99.01 |
| **VGG-MI1 with Augmentation** | **99.02** | **98.76** | **99.78** | **99.49** |
| **VGG-MI1 without Augmentation** | **97.57** | **96.95** | **99.34** | **97.53** |
| **VGG-MI2 with Augmentation** | **99.22** | **99.15** | **99.86** | **99.49** |
| **VGG-MI2 without Augmentation** | **98.07** | **97.95** | **99.60** | **98.53** |

## 6. Discussion

We have observed from Table 9 that the proposed two models after applying transfer learning techniques are shown the best accuracy, sensitivity, predictivity and specificity results when using the first scenario (without noise and with augmentation) comparing to other two models. For the second scenario (without noise and without augmentation), Alex-Net presents the best accuracy, sensitivity, predictivity and specificity results. The VGG-Net shows the worst results in both scenarios compared to other models, while the first proposed model VGG-MI1 shows the worst sensitivity results compared to the proposed second model VGG-MI2 and the other models. In the second scenario, the difference between the VGG-MI1 and the standard VGG-Net in the average sensitivity is 0.3%, which is a small difference, so we can consider that the sensitivity of the VGG-MI1 is acceptable compared to other models. In the first scenario, the discrimination between the VGG-MI1 and Alex-Net is 0.33% and between the VGG-MI2 and Alex-Net is 0.53%. In the second scenario, the discrimination between Alex-Net and the VGG-MI1 is 0.67% and between Alex-Net and the VGG-MI2 is 0.17%. Also, we have observed from Table 9 that there is a big gap in the performance between the proposed two models and the VGG-Net model. The reason for this gap is that we employed two ways of the transfer learning technique, which updated the standard VGG-Net model by reducing the number of pooling layers in our models (we used 3 pooling layers instead of 4 layers). Moreover, in the VGG-MI2 model, we employed QG-MSVM classifier, which plays an important rule for increasing the accuracy of the proposed system. From Tables 7 and 8, we can show that the VGG-MI2 model achieved high accuracy than the VGG-MI1 model. In addition, data augmentation technique plays an important rule for increasing the specificity and the sensitivity rates for all models. Alex-Net gives the best performance in the second scenario (when applying it to a small number of ECG), however, the sensitivity and specificity for the VGG-MI2 in the second scenario are also acceptable comparing to this model. According to this, we intend in the future to apply the transfer learning technique on Alex-Net and observe the results on a small number of ECG segments.

We compared the proposed models with previous MI detection methods on the PTB database as shown in Table 10. Numerous studies have been applied for automated MI detection [10-17,25-27,31,56-60]. Sun *et al.* [15], proposed an automatic system for MI detection. They discussed the rationale for applying multiple instances learning (MIL) to automated ECG classification. They yielded a sensitivity of 92.30% and specificity of 88.10% using KNN ensemble as a classifier. Sharma *et al.* [57], proposed a technique based on multiscale energy and eigenspace (MEES) approach for the detection of MI. This method obtained an accuracy of 96%, a sensitivity of 93% and specificity of 99% using SVM as a classifier and an accuracy of 81%, a sensitivity of 85% and specificity of 77% using KNN as a classifier. Acharya *et al.* [58], proposed a method for automated detection of MI by using ECG signal analysis. They extracted 12 nonlinear features from a discrete wavelet transform (DWT) coefficients. They yielded an accuracy of 98.80%, a sensitivity of 99.45% and specificity of 96.27% using KNN as a classifier. Sadhukhan *et al.* [10], developed an automated ECG analysis algorithm for MI detection. They used harmonic phase values as features, and then they used Threshold-Based classifier for classification. They achieved accuracy, sensitivity and specificity rates of 91.10%, 93.60%, and 89.90% respectively. Sharma *et al.* [60], designed a two-band optimal biorthogonal filter bank (FB) for classification of the MI ECG signals using dataset without noise and noisy dataset. This method obtained an accuracy of 99.74% for the dataset without noise and an accuracy of 99.62% for the dataset with noise using KNN as a classifier. In addition, we updated our previous algorithm [31] to be suitable for MI detection by adding some conditions (according to the cardiologist) in the classification algorithm, such as: If Q-wave is more than 0.04 seconds and more than 25% of the size of the following R-wave the signal is MI. After applying all conditions to the classification algorithm, we achieved an accuracy of 96%, a sensitivity of 95.39% and specificity of 97.22%. Wu *et al.* [25], proposed a deep feature learning-based MI detection and classification approach. They employed SoftMax regressor to perform multiple-class classification with a sensitivity of 99.64% and specificity of 99.82%. Acharya *et al.* [26], implemented a CNN algorithm for the automated detection of MI signals. They used Pan Tompkins algorithm [61] for R-peak detection. They segmented the ECG signals and normalized it using Z-score normalization. Finally, they used 1-dimensional deep learning CNN for training and testing. They achieved the highest accuracy of 95.22% sensitivity of 95.49% and specificity of 94.19% when using noise filters.

Table 10. Summary of a comparative study of the proposed algorithm with selected well-known methods (the results for the proposed approach are given in **bold**)

| Author | Year | Number of ECG beats | Approach | Performances (%) | | |
|---|---|---|---|---|---|---|
| | | | | Accuracy | Sensitivity | Specificity |
| Jayachandran et al. [11] | 2009 | Normal: 2,282 MI: 718 | - DWT - Daubechies 6 wavelet | 96.05 | N/R | N/R |
| Arif et al. [13] | 2010 | Normal: 3,200 MI: 16,960 | - Wavelet transforms - KNN | 98.30 | 99.97 | 99.90 |
| Al-Kindi et al. [14] | 2011 | 40 records | - Digital signal analysis techniques - Clinical background | N/R | 85 | 100 |
| Sun et al. [15] | 2012 | Normal: 79 records MI: 369 records | - Multiple instance learning - KNN ensemble | N/R | 92.3 | 88.1 |
| Chang et al. [16] | 2012 | Normal: 547 MI: 582 | - HMMs - GMMs | 82.5 | 85.71 | 79.82 |
| Liu et al. [17] | 2014 | Normal: 52 records MI: 148 | - Polynomial function - DWT | 94.4 | N/R | N/R |

| | | records | - J48 decision tree | | | |
|---|---|---|---|---|---|---|
| Safdarian et al. [56] | 2014 | 549 records | - Naive Bayes Classification | 94.74 | N/R | N/R |
| Sharma et al. [57] | 2015 | 549 records | - Multiscale wavelet energies<br>- Eigenvalues of multiscale covariance matrices<br>- SVM<br>- KNN | SVM:<br>96<br>KNN:<br>81 | 93<br><br>85 | 99<br><br>77 |
| Wu et al. [25] | 2016 | 10,000 samples | - MDFL | N/R | 99.64 | 99.82 |
| Acharya et al. [58] | 2016 | Normal: 125,652<br>MI: 485,753 | - DWT<br>- KNN | 98.80 | 99.45 | 96.27 |
| Kumar et al. [59] | 2017 | Normal: 10,546<br>MI: 40,182 | - Daubechies 6 (db6) wavelet<br>- FAWT<br>- Sample Entropy<br>- LS-SVM | 99.31 | N/R | N/R |
| Acharya et al. [26] | 2017 | Normal: 10,546<br>MI: 40,182 | CNN | With noise:<br>93.53<br>Without noise<br>95.22 | 93.71<br><br>95.49 | 92.83<br><br>94.19 |
| Hammad et al. [31] | 2018 | 549 records | - Characteristics of ECG signals | 96 | 95.39 | 97.22 |
| Dohare et al. [12] | 2018 | 120 records | - Principal Component Analysis<br>- SVM | SVM:<br>98.33<br>SVM with PCA:<br>96.96 | 96.66<br><br>96.96 | 100<br><br>96.96 |
| Sharma et al. [60] | 2018 | Normal: 10,546<br>MI: 40,182 | - Two-band optimal biorthogonal filter bank<br>- KNN | Without noise:<br>99.74<br>With noise:<br>99.62 | 99.84<br><br>99.76 | 99.35<br><br>99.12 |
| Liu et al. | 2018 | MI: 13,577 | CNN | With noise: | | |

| [27] | | Normal: 3,135 | | 98.59 | 99.79 | 94.50 |
| --- | --- | --- | --- | --- | --- | --- |
| | | | | Without noise: | | |
| | | | | 99.34 | 99.79 | 97.44 |
| Sadhukhan et al. [10] | 2018 | MI: 15,000 Normal: 5,000 | - Harmonic phase distribution pattern<br>- Threshold-Based Classifier<br>- Logistic Regression (LR) | Threshold:<br>91.1<br>LR:<br>95.6 | 93.6<br><br>96.5 | 89.9<br><br>92.7 |
| **Proposed** | **2019** | **Normal: 21,092 MI: 80,364** | - CNN<br>- QG-MSVM | **VGG-MI1**<br>**99.02**<br>**VGG-MI2**<br>**99.22** | **98.79**<br><br>**99.15** | **99.49**<br><br>**99.49** |

**\* N/R: Not Reported.**

**\* DWT: Discrete Wavelet Transform; KNN: K-Nearest Neighbor; HMMs: hidden Markov models; GMMs: Gaussian mixture models; SVM: Support Vector Machine; DFL: deep feature learning; FAWT: flexible analytic wavelet transform; LS-SVM: least-squares support vector machine; CNN: Convolution Neural Network; QG-MSVM: Q-Gaussian multi-class support vector machine.**

It is evident from the analysis of the results that our proposed algorithm is more robust compared to other works that mentioned in Table 10. Most of the previous researches [14-17,56,57] work on a small number of ECG segments, which gives a low sensitivity rate. In our study, we solved this problem by using data augmentation technique, which gives high specificity and sensitivity rates. Furthermore, the methods proposed in [10,12-17,25,57,58] used ECG recordings of more than one lead, however, we used only lead II ECG signals, which makes our methods less complex than other methods that used more than one lead. Moreover, our methods achieved the highest accuracy comparing to the existing methods that are used CNN and mentioned in Table 10. In addition, we updated our previous work [31] to detect the MI signals, but it gives lower accuracy (96%) comparing to the proposed method. The reason for the low accuracy of our previous work is that this method is sensitive to the ECG signal quality and cannot detect the noisy MI signals correctly. However, the proposed method performed well for clear and noisy MI signals.

The advantages of our proposed algorithm are summarized below:

- Transfer learning is successfully used to increase the robustness of the proposed models.
- The proposed system achieves superior results compared with the previous systems based on deep learning approach.
- Ten-fold cross-validation strategy is used in this work. Hence, the results are robust.

- Data augmentation plays an important rule to increase the robustness of the proposed system against small variations.
- Segmentation and feature extraction are no longer required in the proposed method.
- Using small filter size in the first convolution layers and a small number of pooling layers lead to lower computation cost and make the proposed models more stable when using the input ECG image.
- Using QG-MSVM classifier lead to solve the small sample problem that faces the CNN training, which usually requires a large sample data.
- The proposed algorithm can help the experts to detect the MI signals more precisely and considered deploying in hospitals and clinics.

The disadvantages of the proposed work are as follows:

- The proposed algorithm is costly compute comparing to machine learning algorithms.
- Need to test more data using the proposed model.

## 7. Conclusion and Future work

In this work, we have proposed an effective system to detect MI signals for urban healthcare in smart cities. The new approach used the *two*-dimensional CNN. We have employed two ways of the transfer learning technique to retrain the pre-trained VGG-Net and obtained two new networks VGG-MI1 and VGG-MI2. Moreover, data augmentation techniques are employed to increase the classification performance. We have utilized two-seconds ECG signals from the PTB database to evaluate the effectiveness of our proposed method according to four different scenarios. Experimental results show that the best result is obtained when using the first scenario (best-case1) on the second model which is more accurate and robust compared to other previous works. We have achieved an accuracy of 99.22%, a sensitivity of 99.15% and specificity of 99.49% when using the best scenario on VGG-MI2. Also, we have achieved an accuracy of 97.24%, a sensitivity of 97.13% and specificity of 96.53% when using the worst scenario on VGG-MI2. This suggests that the proposed method can detect the MI signals with high-performance results even though there are noise present in the ECG beats. Hence, it is obvious that the proposed algorithm has the possibility to accurately diagnose MI signals and considered deploying in hospitals and clinics.

In the future, we intend to extend our method to detect different types of heart diseases such as atrial fibrillation (A-Fib), ventricular fibrillation (V-Fib) and atrial flutter (AF). Also, we intend to use more augmentation types to improve the performance of our method. In addition, we attend to apply the proposed algorithm on different physiological signals such as electroencephalogram (EEG) [63,64] and observe the effect of using our model on the EEG performance. Finally, we attend to apply other models on the MI data and observe the effect of these models on MI detection results.

**Acknowledgment**

Ahmed A. Abd El-Latif acknowledges support from TYSP-Talented Young Scientist Program (China) and Menoufia University (Egypt). Additionally, the authors warmly thank their families for their unconditional support.